\documentclass[sigconf,nonacm]{acmart}
\AtBeginDocument{%
  \providecommand\BibTeX{{%
    \normalfont B\kern-0.5em{\scshape i\kern-0.25em b}\kern-0.8em\TeX}}}

\setcopyright{acmcopyright}
\copyrightyear{2018}
\acmYear{2018}
\acmDOI{XXXXXXX.XXXXXXX}

\acmConference[Conference acronym 'XX]{Make sure to enter the correct
  conference title from your rights confirmation emai}{June 03--05,
  2018}{Woodstock, NY}
\acmPrice{15.00}
\acmISBN{978-1-4503-XXXX-X/18/06}

\begin{document}

\title{Automated Argument Generation from Legal Facts}

 \author{Oscar Tuvey}
\email{hsotuvey@liverpool.ac.uk}
\affiliation{%
   \institution{University of Liverpool}
   \streetaddress{P.O. Box 1212}
  \city{Liverpool}
   \country{United kingdom}
}
 \author{Procheta Sen}
 \email{procheta.sen@liverpool.ac.uk}
 \affiliation{%
   \institution{University of Liverpool}
   \streetaddress{P.O. Box 1212}
  \city{Liverpool}
   \country{United kingdom}
}








\renewcommand{\shortauthors}{}

\begin{abstract}
The count of pending cases has shown an exponential rise across nations (e.g., with more than $10$ million\footnote{\url{https://njdg.ecourts.gov.in/njdgnew/index.php}}  pending cases in India alone). The main issue lies in the fact that the number of cases submitted to the law system is far greater than the available number of legal professionals present in a country. Given this worldwide context, the utilization of AI technology has gained paramount importance to enhance the efficiency and speed of legal procedures. In this study we partcularly focus on helping legal professionals in the process of analyzing a legal case. Our specific investigation delves into harnessing the generative capabilities of open-sourced large language models to create arguments derived from the facts present in legal cases. Experimental results show that the generated arguments from the best performing method have on average $63\%$ overlap with the benchmark set gold standard annotations\footnote{This paper has been accepted as a non-archival submission in MLLD workshop in CIKM 2023}. 
\end{abstract}






\maketitle

\section{Introduction}\label{sec:intro}
Given the recent surge in pending cases worldwide, the utilization of AI tools to efficiently process legal documents has grown more imperative. Legal documents are usually long and complicated, with lots of information in them \cite{unknown}. This makes it take a long time for lawyers to read them thoroughly. Creating arguments for cases depends on lawyers understanding the whole document very well. But this task takes a lot of time, especially with so many cases that are still waiting to be solved.

To address this challenge, we propose a supervised model aimed at automatically suggesting arguments based on the factual content within legal cases. Our approach centers on employing large language models to automatically generate arguments for legal cases. These arguments can greatly aid legal practitioners in preparing for cases more efficiently.
In our research scope we specifically focused on open-sourced  Large language models (LLMs). LLMs typically require substantial training data. To the best of our knowledge, there is no publicly accessible datasets featuring annotations for both facts and corresponding case arguments. The work in \cite{argument_extract} proposed a supervised model to automatically annotate seven rhetorical labels from legal documents in India. Consequently, we use the model proposed in \cite{argument_extract} to extract facts and arguments required for training our model. In our research scope, we particularly focus on legal case proceedings from India. 

Our experimentation encompasses the utilization of LLMs like FLan-T5 \cite{chung2022scaling} and GPT-2 \cite{radford2019language}. One of the major challenges faced in using GPT-2 and FLan-T5 was their \textit{max token} limitation. The facts or arguments for a legal case were in general longer than \textit{max token}. As a result of this, we leveraged summarization technique to generate facts and arguments of reasonable size. We evaluate the generated arguments against a set of manually (i.e. by legal experts) annotated arguments. Our findings show that FLAN-T5 outperforms GPT-2 in argument generation.

\section{Related Work}
Existing work related to our research scope can be broadly categorized into two different areas. They are Legal AI and large language models. Each one of then is described as follows.
\subsection{Legal AI}
In legal cases, the documents often encompass lengthy and intricate sentences, making it challenging and time-consuming to thoroughly read and comprehend the entire content of a case document. To alleviate this extensive effort, researchers have emphasized the extraction of noun phrases known as \textit{catchphrases }\cite{catchphrase1, catchphrase2} from the document. This approach aims to capture the key elements and central themes of the text. Additionally, \textit{summarization} \cite{ANAND20222141,shukla-etal-2022-legal} techniques have been proven to be effective in gaining a comprehensive understanding of the document by condensing its content into a concise summary. In terms of down stream application there has been extensive work on judgement prediction \cite{Zhang}, statute prediction \cite{pmlr-v101-feng19a} using advanced NLP approaches. To the best of our know ledge there has been no work on generating arguments from legal facts. 

\textbf{Reasoning in AI and Law}
Early attempts at imitating legal reasoning were made in 1976 by McCarty's TAXMAN \cite{TAXMAN}, which aimed to replicate arguments in the notable tax law case \textit{Eisner v Macomber}. Similar efforts were made to the US Trade Secrets Domain by HYPO, which pioneered a rule based approach for emulating legal reasoning \cite{HYPO}. This methodology was explored by subsequent others \cite{HYPOlegacy}, in particular by CATO \cite{CATO}. CATO popularized the notion of \textit{factors}: typical factual patterns that support either the plaintiff or defendant's perspective \cite{mumford2021machine}. 
The study in \cite{PredictingDecisions} provided a list of 20 words listed in order of their significance on the case outcome. However, the lists included words which implicated that the algorithm was relying upon elements of the data which had no significance on the actual case outcome (such as months of the year), indicting a limited grasp on legal reasoning. The study in \cite{Prakken} provided a means by which legal judgement prediction systems could explain outputs using factors. However, this assumes that factors in of themselves can be used to justify the case outcome; as argued in \cite{bench2021precedential}, factors should be used to justify resolution of issues, which in turn justify the case outcome, and not the outcome of the case itself.

\subsection{Large Language Model Application in NLP}\label{sec:relWork}
In recent literature Large Language Models (LLMs) have been successfully applied in a range of Natural Language Processing tasks like Machine Translation \cite{zhang2023prompting}, Summarisation \cite{liu2023learning}, Entity Recognition \cite{unknown}. The inception of LLMs started with Transformer architecture\cite{NIPS2017_3f5ee243}. Transformer architecture mostly used attention mechanism and was successful in generalizing most of the tasks. Broadly speaking, LLMs are used in three different ways in NLP tasks a) Pre-trained models, b) Fine-Tuned models and c) Prompt based application. Pre-trained models are trained on a large amount of unannotated data through self-supervised training and can be applied on any kind of NLP task. Pre-trained models like BERT \cite{bert}, GPT-2 \cite{radford2019language} have over performed state-of-the-art baselines in a range of NLP tasks \cite{WANG2022}. Fine-tuned models \cite{xu-etal-2021-raise} are trained on a particular task and while fine-tuning the model is initialized with the pre-trained parameters. Existing research has shown that LLMs fine-tuned on specific downstream tasks performs better than training the model from scratch on the downstream tasks \cite{jensen-plank-2022-fine}. With recent release of GPT 3.5 and GPT-4 prompt base learning have also gained attention. In prompt learning \cite{prompt}, frozen LLMs are used with task specific prompts (i.e. set of keywords or tokens) to get better outputs for that particular task. Prompt based learning is less memory consuming than training a model from scratch or fine tuning. The study in \cite{Peric2020LegalLM} first demonstrates the potential of large language models for generating legal text. In this work, we used only fine-tuning to generate arguments from legal facts.

\section{Argument Generation Framework}\label{sec:annotate}
As described in Section \ref{sec:intro}, the first step for argument generation is to automatically extract facts and arguments from legal documents. We follow the approach proposed in \cite{argument_extract} for extracting facts and arguments. The details of the process is described in the following subsection.

 \paragraph{\textbf{Extraction of Rhetorical Roles}} \label{argument:extract}
The study in \cite{argument_extract} showed that a legal case can be broken down into several rhetorical roles. The different labels are a)Facts, b) Ruling by Lower Court, c) Argument, d) Statute, e) Precedent, f) Ratio of decision, g) Ruling by Present Court. Facts refer to the chronology of events that led to the filing of the case, and how the case evolved over time in the legal system. Ruling by Lower Court  refer to the judgments given by lower courts (Trial Court, High Court). Argument refers to the discussion on the law that is applicable to the set of proven facts. Statute refers to the established laws, which can come from a mixture of sources. Precedent refers to the prior case documents. Ratio of decision refers to the application of the law along with reasoning/rationale on the points argued in the case. Ruling by Present Court refers to the ultimate decision and conclusion of the Court.

The work \cite{argument_extract} used a Hierarchical BiLSTM CRF model \cite{huang2015bidirectional} to automatically assign one of the seven rhetorical labels mentioned above to each sentence of a legal case document. They used a manually labelled data by legal experts to train the BiLSTM CRF model. In the context of this research, we are only interested in sentences belonging to `Facts' and `Ratio of Decision' category in a legal document. Please note that the `Ratio of Decision' has the reasoning and rationale in contrast to the `Argument' label. Consequently we choose `Ratio of Decision' for argument generation problem setup.

\paragraph{\textbf{Generation Methodology}}
We used GPT-2 \cite{radford2019language} and Flan-T5 \cite{chung2022scaling} models for argument generation. The generative model takes facts as input and produces arguments as output. Broadly speaking, LLMs learn from the training data by learning to predict the next word. As described in Section \ref{sec:relWork}, that LLMs are generally used in three different ways. In our approach we used fine tuning to generate arguments corresponding to the facts. Special tokens are inserted in the training set to give instructions to the LLMs. In our research scope, we used special tokens like `[Facts]' and `[Arguments]' for fine tuning LLMs. 

As described in Section \ref{sec:intro}, legal documents are in general long. Consequently the `Facts' and `Arguments' obtained from the rhetorical model labeling models are in general long. However, the \textit{max token} length for text generation for GPT-2 and FLAN-T5 are $1024$ and $512$ respectively. As a result of this we have used BERT-summarizer\footnote{\url{https://pypi.org/project/bert-extractive-summarizer/}} to summarize the facts and arguments corresponding to a legal case. Summarized facts and arguments are within the \textit{max token} range of LLMs.

\section{Experiment Framework}
\paragraph{Dataset}
We used a manually annotated dataset of 50 documents released in \cite{argument_extract} from the Indian Supreme Court corpus\footnote{\url{http://www.liiofindia.org/in/cases/cen/INSC/}} as our benchmark dataset. The datset in \cite{argument_extract} assigned exactly one of seven rhetorical role labels: facts, ruling by lower court, argument, statute, precedent, ratio of the decision, and ruling by present court \cite{rhetoricalroles} to each sentence. The rhetorical role label was based on the majority verdict of the three annotators, who were all senior Law students. To add more data in the training set we randomly chose another $50$ documents from the Indian Supreme Court corpus which consists of $6,560$ case proceedings. Out of $50$ gold standard documents $40\%$ of the randomly chosen data (i.e.  $20$ documents) was used as test data in our experiment setup. The details of the dataset statistics is reported in Table \ref{tab:ubar}.

A significant concern regarding the data presented in Table \ref{tab:ubar} pertains to its high level of noise, characterized by a substantial portion of poorly structured English sentences. To tackle this problem, we conducted a comprehensive re-writing of all the samples using GPT-3.5. This procedure retained the original sentences' intended meanings while ensuring they were now properly structured and grammatically correct. Subsequently, we performed meticulous manual verification on each of the newly created samples to ascertain the preservation of all the information from the original dataset. Our hypothesis was that the use of well-structured English sentences would lead to improved performance for any language model.



\begin{table}[H]
    \footnotesize
    \centering
    \begin{tabular}{|l|c|c|c|}
    \hline
    & \multicolumn{3}{|c|}{\textbf{Statistics}} \\
    \hline
    \textbf{Data Type} & \textbf{\#Docs} & \textbf{\#Avg Words} & \textbf{\#Avg Sentences} \\
    \hline
    Train Data & 70 & 2597.8 & 148.1 \\
    Test Data & 20 & 3328.2 & 123.3 \\
    Validation Data & 10 & 3149.1 & 262.9 \\
    \hline
    \end{tabular}
    \caption{Automated Tag Extraction Results using Flair}
    \label{tab:ubar}
\end{table}
We investigated two different variations of summaries for both GPT-2 and FLAN-T5 using BERT-Summarizer. They are $3$ sentence and $5$ sentence summaries. The average length of arguments in $3$ sentence and $5$ sentence summaries are $90.12$ and $120.35$ respectively. For evaluation purpose we have mainly used two different evaluation metrics. They are Average Word Overlap and Average Semantic Similarity. Average word overlap computes the average number of words common between the ground truth arguments and the generated arguments.  Average Semantic similarity computes the cosine similarity between the BERT embeddings corresponding to the ground truth argument and the generated argument.

 \section{Results}
Table \ref{tab:ubar} shows the quality of the generated arguments for different variations of LLMs. Flan-T5 using 5 sentence summaries have performed the best among all the approaches which uses the original data. There are mainly two observations from the Table reported in Table \ref{tab:ubar}. Firstly, it can be observed that with increase in the number of sentences in the summary, the quality of the generated argument also increased. The second observation is that with a better quality dataset the performance significantly improves for the same model (i.e. as observed in the last two rows of the Table \ref{tab:ubar}).
 \begin{table}[H]
    \footnotesize
    \centering
    \begin{tabular}{|l|c|c|c|c|c|}
    \hline
  &\multicolumn{2}{|c|}{\textbf{Data Format}}  & \multicolumn{2}{|c|}{\textbf{Evaluation Metric}} \\
    \hline
    \textbf{LLM} &\textbf{\#Sent} &\textbf{Source} & \textbf{Avg Word Overlap}  & \textbf{Avg Semantic Sim} \\
    \hline
    GPT-2 &3 &Original &15.12\%  & 0.335 \\
     GPT-2 &5 & Original&16.31\%  & 0.340 \\
      FLAN-T5 &3  & Original& 32.41\% & 0.376 \\
   FLAN-T5 &5  & Original&31.31\%& 0.387 \\
   \hline
      FLAN-T5 &5 &GPT 3.5& \textbf{63.13\%}  & \textbf{0.492} \\
    \hline
    \end{tabular}
    \caption{Quantitative Evaluation of Generated Arguments using Different LLMs.}
    \label{tab:ubar}
\end{table}

\section{Conclusion}
In this paper we did an initial exploration to investigate the effectiveness of the generative models in generating arguments given the facts of the legal case. In future, we would like to explore advanced prompt tuning approaches to generate better quality legal arguments from facts.
\bibliographystyle{acm-reference-format}
\bibliography{main}
\end{document}